ANOMALY DETECTION: Review and
preliminary Entropy method tests.

Pre-Dissertation Project in Signal Processing and
Communications (MSc) - PGEE11109

OLUWASANYA, PELUMI WONUOLA
(s1423789)

**PROJECT MISSION STATEMENT**

**Project Title: Anomaly Detection in Wireless Sensor Networks**

Supervisors: Professor Bernard MULGREW

Student's Name: Pelumi OLUWASANYA

PROJECT DEFINITION

A primary motivation for use of wireless sensor networks is change detection over large areas. This might be environmental change such as temperature of the atmosphere or the presence of some characteristic in the received signal that is different from what has been received before. This is anomaly detection, which is, significantly more challenging than conventional detection where we know the signal we wish to detect. Many methods have been proposed for anomaly detection. The one that will be explored in this project is based on estimating the entropy of a signal directly from the data. The entropy is itself estimated through first estimating the probability density function using the k-nearest neighbour (k-NN) technique. This method will be applied to actual measured data.

1) *The pre-dissertation phase will involve a literature survey on the use of anomaly detection within wireless sensor networks as well as preparing a Matlab simulation of suitable test signals*;
2) Simulation and testing of the k-NN technique of [1] on the synthetic test signal evaluated in step 1 to evaluate performance;
3) Verification of the results of [1] on the recorded signals found at: www-personal.umich.edu/~kksreddy/rssdata.html;
4) If time permits, modification of the algorithm to improve performance.

Supervisor: _______________________________

Student: _______________________________

Date: ____________

**ABSTRACT**

Anomalies are strange data points; they usually represent an unusual occurrence. Anomaly detection is presented from the perspective of Wireless sensor networks. Different approaches have been taken in the past, as we will see, not only to identify outliers, but also to establish the statistical properties of the different methods. The usual goal is to show that the approach is asymptotically efficient and that the metric used is unbiased or maybe biased.

This project is based on a work done by [1]. The approach is based on the principle that the entropy of the data is increased when an anomalous data point is measured. The entropy of the data set is thus to be estimated.

In this report however, preliminary efforts at confirming the results of [1] is presented. To estimate the entropy of the dataset, since no parametric form is assumed, the probability density function of the data set is first estimated using data split method. This estimated pdf value is then plugged-in to the entropy estimation formula to estimate the entropy of the dataset.

The data (test signal) used in this report is Gaussian distributed with zero mean and variance 4. Results of pdf estimation using the k-nearest neighbour method using the entire dataset, and a data-split method are presented and compared based on how well they approximate the probability density function of a Gaussian with similar mean and variance. The number of nearest neighbours chosen for the purpose of this report is 8. This is arbitrary, but is reasonable since the number of anomalies introduced is expected to be less than this upon data-split. The data-split method is preferred and rightly so.

Also presented in this report is a critical literature review of the works that have been done in Anomaly detection in wireless sensor networks in the past few years, identifying their strengths and weaknesses as well as comparing and contrasting them. Also identified are the challenges of anomaly detection and the requirements that have now thus been formed.

Finally, this report offers some projection into what the main project phase would seek to add as well as present a Gantt chart that shows the project plan, expected challenges as well as possible solutions.

**TABLE OF CONTENTS**

**List of Acronyms**

k-nn, k-NN    –    k –nearest neighbours

pdf           –    probability density function

K-knn         –    K- k –nearest neighbours

ROC           –    Receiver Operating Characteristic

RSS           –    Received Signal Strength

Q-Q Plot      –    Quartile - Quartile Plot

DWT           –    Discrete Wavelets Transform

SOM           –    Self-Organizing Maps

MAD           –    Median absolute deviation

**List of Figures**

# 1.0 INTRODUCTION

## 1.1 Sensors, an overview

Sensors are transducers [2]. They convert a physical quantity into electric signals. They are used to monitor events and activities in their environment. Sensors usually send these electric signals to some location for processing and monitoring. There are different types of sensors. They are usually classed based on what they sense. When these sensors are connected to this location through a wireless link, they are said to be wireless sensors, when this link is through a wire, they are called wired sensors. A large number of sensors interconnected in any of these ways form a network.

Sensor networks are very important to our everyday lives. From traffic control, to intrusion detection, to agricultural monitoring, security, healthcare etc. Sensor networks let us know the state of our surroundings by reporting changes in the value of certain variable(s) they measure. These measured changes are useful for decision-making, diagnostics, etc. The enormity of the decisions that can be taken based on sensor measurements makes it imperative that the integrity of data measured by sensors must never be in doubt. It is essential to have sufficient level of confidence in the measured data such that the changes in the values measured can be securely assumed to represent a change in the value of the variable being measured, and also to be made aware of the amount of confidence that can be safely placed on the measured data as returned by the sensors. Sadly, this is not always so.

A point in this network where more than one sensor meet is called a node [3]. Sensor nodes may vary in size depending on the location of each of the sensors connected to it or the exact application requirements.

Both types of sensor networks may be preferable in different circumstances. A wireless sensor network is usually preferred on the grounds of mobility and flexibility. Changes in location can be easily achieved with this type of network of sensors. Environments that require constant movement, such as the human body, must almost always use this type, as a mesh of wires may not be desirable to carry around when human physiological data is being monitored in out-of-hospital environments. However, environments with very strict accuracy requirements in measurements and very short distances e.g., patient health monitoring in a hospital ward, will prefer a wired network as this will, in spite of its own internal noise, still do better than a wireless sensor in this condition.

Processing of data obtained in a sensor network may be done at the node level or at a central location in the network. The former is called a distributed network while the latter is called a centralized network. Both methods have their advantages and drawbacks and are both preferred in different situations depending on the particular network structure. Small size networks where sensors are located not very far from each other are usually use of the centralised system. Whereas larger networks and where sensors are located at quite some significantly large distance from each other or form clusters about different points in the network that are arbitrarily far from each other would prefer the distributed approach as this helps to reduce communication overhead and power consumption in the network.

Sensor thermal characteristics, internal faults, and temperature drifts [1] or even malicious attacks [4] can seriously degrade the performance, if not totally render useless or unrepresentative of the ambient conditions of the environment, the data reported by the sensors. Wireless sensor networks are particularly prone to all manner of losses either from objects in their surroundings or attenuation of signals with distance.

In many situations where wireless sensors are used, there is usually some sense of a set of expected values to be returned by the sensors. A reasonable example is a sensor network monitoring the amount of sunlight in a farm. Previous measurements give a range of values that the temperature should lie in. A measurement that does not lie within this expected region is known as an anomaly.

Anomalies are also called outliers [5]. In a set of measurements known to be without anomalies, there is an assumed underlying distribution model; this may be more than one in many cases. An anomalous data point is known as a data point that exhibits behaviours that makes one think that it must be the result of a different distribution model compared to other data points within the same dataset [6]. The approach would be to identify the underlying distribution models and mark one that does not follow these as anomalous. Identifying these models is probably the most challenging task in anomaly detection.

From the foregoing, anomaly detection is essential but inherently difficult [7]. In certain instances, an anomaly can be easily detected, in many other cases this is not so. This occurs because the very fact that it is an anomaly makes it difficult to predict what form it may take.

The general approach however, is to have a clear and correct knowledge of what is the normal situation. Hence, the underlying distributions or actual/most important (defining) variable characteristic(s) are determined or specified, and any data point that does not correspond to this distribution or exhibit this characteristic(s) is classed as an anomaly, usually with some confidence level or false alarm rate [8].

This project focuses on wireless sensor networks since the goal is to detect anomalies over large areas in which a large number of sensors are deployed. Some approaches to this detection use the distributed method while some use a centralised method. These will be seen in literature review presented in chapter two. Also monitoring changes in variables like temperature or pressure over a large area presents some knowledge or expectation of how quickly these parameters can change.

The main goals of the project are presented below in terms of aim and objectives:

**1.2 Aim**

1. To detect anomalous data points in a data set with a confidence level of 95% and a 0.05% false alarm rate.

**1.3 Objectives:**

1. Estimate the probability density function (pdf) of the data by splitting the dataset into two arbitrary parts.
2. Estimate the entropy of the dataset using the pdf evaluated using the other data points in the other subset.
3. Set a threshold for detection
4. Determine the confidence interval.
5. Verify statistical properties via receiver operating characteristics (ROC) plots and Quartile-Quartile (Q-Q) plots.

## 2.0 LITERATURE REVIEW
## 2.1 Background

Anomaly detection in wireless sensor networks has been studied quite intensely [6] [9] though approach has been from different directions and applications have been just as divergent. Researchers have been concerned with the several areas of challenge. A very popular one is the curse of dimensionality [10]. The performance of many anomaly detection methods degrade tremendously as the dimensions increase. It is usually no longer cheap to compute the parameters required to arrive at the results presented. Also, most important parameters of all measured data, especially when in the presence of noise are believed to lie in a subspace of lower dimensions, hence it is also beneficial many times to map the data first into such subspace. Another challenge is however created, which is, determining the right subspace. Techniques like the principal component analysis [11], etc., have been used to reduce the dimension of observation data and many times as a pre-processing stage before the anomaly detection algorithm is applied.

Communication cost is another important issue especially in large networks where many sensor nodes are connected together. This has resulted in the use of distributed processing where some processing is carried out at the sensor nodes [7] and centralized in much smaller networks where the cost of communication is not excessive.

Interpretability is also a major issue. This was briefly mentioned in the introduction. It is arguably the most important duty of an anomaly detection system i.e. to be able to recommend some diagnostic action and not just to spot the irregularity. In this sense, the challenge is to be able to identify in each case the causes of the anomalies, their peculiar characteristics that can be used to differentiate them. For example, in [12], anomalous measurements are due to either an error or an event. Sensor faults and equipment intrinsic characteristics, etc., are grouped together as errors while the actual variation in the variable being monitored is regarded as an event. It is essential to be able to distinguish between the two from the data obtained [13]. In some other cases such as [14], anomalies are said to also be due to faulty communication; in addition to the two presented in [12]. There must be a way to effectively interpret all the possible measurements in each method with low false alarm and missed detection rates and specify intuitively helpful solutions.

Furthermore, it may be desired to have a detection system that is both online and adaptive. In applications to healthcare [15], a sudden deterioration in the health of a patient must be reported to the healthcare practitioners immediately for prompt action and continuous changes must be monitored in order to save such lives. Nothing less than that is acceptable. Systems designed must also be able to adapt with minimal change requirements to different environments. These and all the others have given rise to what is generally known today as the requirements for anomaly detection in wireless sensor networks [16] and different methods proposed so far usually settle for some trade-offs in order to deliver the best possible performance of that method.

## 2.2 Related Works

Anomaly detection methods are divided into the following groups [13] depending on the metric used: extreme value-based, proximity-based, probabilistic/statistical methods, information theoretic methods, etc.

The first method assumes that in a measured dataset, anomalies usually lie at the very edges of the dataset. So values that lie at the ends of a set of measured will be flagged as anomalies depending on the threshold of the detector. While this is logical, a second thought shows that this detecting system implicitly assumes that the entire dataset can be accurately described by only one cluster and so can deliver good results only when this is true. Unfortunately, this is not always true in real life scenarios: in situations where more than one cluster exists, the results are poor. To see this more clearly, consider the dataset {1, 4, 4, 4, 5, 5, 4, 5, 5, 10}, an extreme value analysis detector would return {1, 10} as the anomalous data points, and rightly so. But consider the following dataset with two clusters {1,2,1,2,1,3,6,9,10,10,10} extreme value analysis method fails to provide desirable result in this case. Also in the multi-dimensional case, spikes in the midst of a dataset cannot be identified by extreme value analysis. However this method is computationally efficient since the most important calculation to be carried out is to determine the range of the distribution and to step off some values at the edges of the dataset.

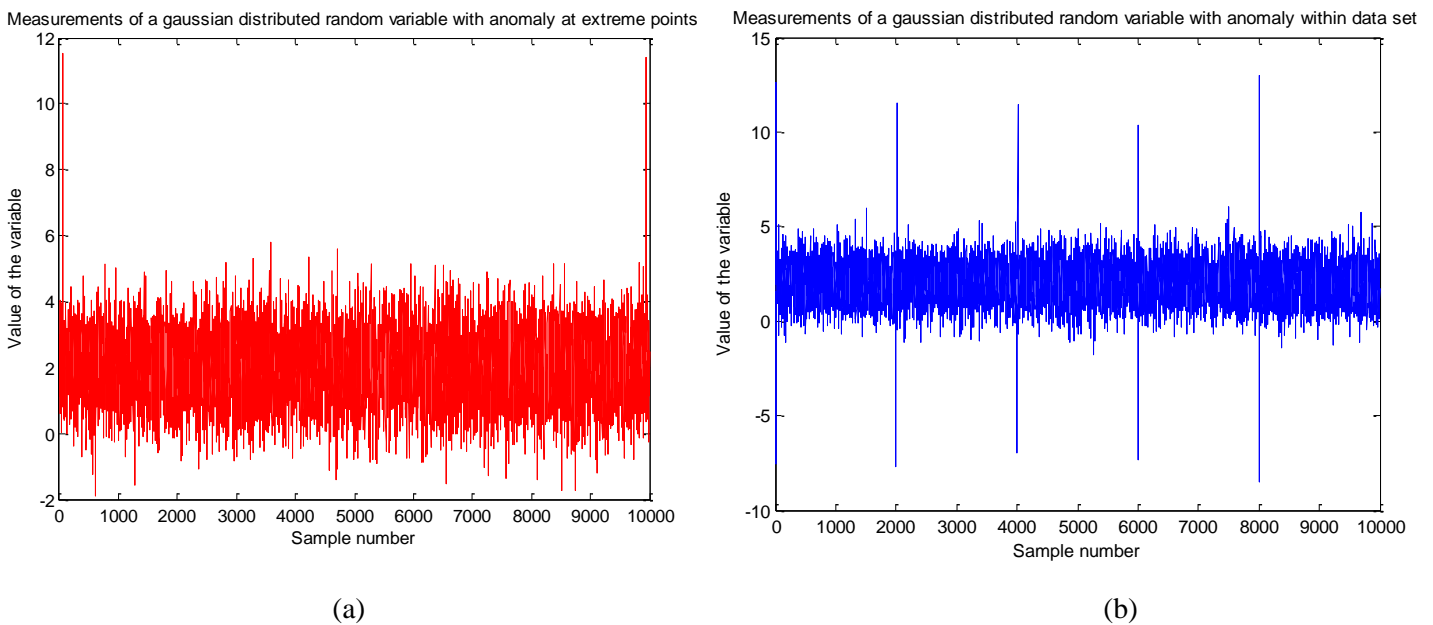

*Figure 2.1:* Different positions of anomalies in a dataset (a) extreme anomalies (b) anomalies within the dataset

Proximity based methods rely essentially on the fact that an outlier would usually lie farther to all other data points than all other data points. Some measure of distance is used and any data point with distance value greater than a particular threshold is declared an anomaly.

Depending on the variant, proximity-based methods can detect anomalies in a multiple cluster situations since the distance measure is determined for all the data points and compared with others e.g. the nearest neighbour method and are regarded as being more granular [13]. This method usually doesn't assume a probability distribution model for the dataset. This ensures that the designed can easily be adapted to any data. However, the method is largely computationally more expensive. Clustering based method is a variant of the proximity-based method in that it relies on some measure of distance from other data point. Here however, data is assumed to cluster about some central values called the mean. Usually there are k-means in the normal data set; distances to these k means are measured for all the data points. They are then allocated to their respective groups based on this. However, a threshold is usually set for how far any data point can be from its closest mean. Any data point with values beyond this limit is flagged as an anomaly. [17] Presents one such method. It however seeks to improve the algorithm by using a measure of the entropy. This is used iteratively to correct the result of the k-means algorithm.

Statistical methods are divided into parametric or non-parametric based on whether there is an assumed distribution for the data before analysis or not. Of course both have their shortcomings. The parametric methods simplify calculations and reduce the problem of determining the distribution to that of estimating a few parameters and they also provide very good results if the data follows the distribution assumed but however may fail woefully if the data does not follow the assumed model. Non-parametric methods are usually more computationally expensive but will usually be guaranteed to provide a fair result depending on the actual approach taken. So the decision about which approach to take is usually dependent on the particular application in question. This is because while the proponents of non-parametric methods guarantee good results always, those of parametric methods especially using mixture of Gaussians or just Gaussians argue that according to the central limit theory, if your dataset is large enough, the result would always tend to be Gaussian distributed [8], thus assuming some form of Gaussian distribution or a mixture of Gaussians would always closely if not completely match the data distribution. A statistics-based anomaly detection system is presented in [12]. It takes advantage of correlation in measured data from same location (spatial) at different times (temporal) with equal time lags. It however, uses auto-regressive moving average to model a time correlation and values obtained that are outside a range from the predicted value are flagged as anomalies. This also applies to the spatial correlation in the spatial domain.

The approach in [15], is from a seemingly non-parametric point of view, even though the authors still use the Gaussian to validate their results, it is claimed that the method will work with no prior information about the signal. It is an online method. Anomaly detection is applied to a medical wireless sensor network comprising sensors placed on different parts of the body of a patient to take physiological readings, and the problem is modelled such that the proposed system saves network resources by only transmitting data only when a significant event occurs. It also reduces calculation complexity by not using the mahalanobis distance, which requires the calculation of an inverse of a matrix. Sensors in this method use an exponentially weighted moving average to predict the next data point to be measured by the sensor. This predicted value is then compared to the actual measured value. An enclosing ellipse is then calculated based on the values of the actual measurement and the error. The error is expected to be very small for normal measurements. Since it is also expected that the detection system will only provide an alert when a true change occurs in the state of health of the patient being monitored, and knowing that physiological measurements are correlated, a threshold of two sensors was set in the work as the number of sensors returning anomalous values for further processing to be done. Finally a kernel density estimator is used on the windowed Chi-Square distance of the errors and an anomalous measurement is finally detected as one, which has a pdf different from the others. This method has a disadvantage of detecting anomalies only when there is significant deviation from the norm, and therefore only spots the edges of change. For example, if in a hospital, a patient is seriously in trouble, this anomaly detection system will only return detection at the point of deterioration and then does not flag an anomaly until when the state of health of the patient improves significantly or deteriorates further. This may be undesirable especially when constant attention needs to be called for attention to the patient while sick, in this case, it might be very much desirable to have the system flagging anomaly when in an unpleasant state until the patient recovers. This would be a constant feedback to healthcare officials working to take care of the patient. The system was tested using medical records for heart rate, pulse, etc., and it performed quite well as reported by the authors.

[14] Presents an application of anomaly detection to sensors in an agricultural environment using discrete wavelet transforms (DWT) and self-organising maps (SOM). The sensors in this system monitor many parameters at once and these are grouped together as a state vector. Firstly a DWT is performed to split the measured values into the detail coefficients and the approximation coefficients. The detail coefficients represent more of the underlying features of the distribution of the data and are used to train the system while the approximate coefficients are then used to test the system. The SOM uses the distance between any state vector and the winning neuron vector to flag an anomaly. A threshold is set and if exceeded, as usual, anomaly is flagged. Else, the dataset is accepted as normal. This method tries to model the detector based on the actual subspace that contains the important properties of the system by using the detail coefficients of the discrete wavelet transform so it can be assured to produce desirable results for measurements of the same parameter. The authors have presented simulation results for up to 99% confidence interval and significantly lower false alarm

rates compared with other methods used. However the system is quite computationally expensive to train, though much cheaper to use afterwards. Also, since the difference in distance between two, potentially high dimensional vectors is considered, a lesser difference in all the vector elements between two measured state vectors, and a massive difference between a few elements of the two of such vectors would return similar results. This is quite misleading, and an anomaly may be wrongly called or missed depending on the threshold. The authors refer to a local decision to circumvent this problem but this is only done after an anomaly has been flagged.  Also, it is not certain whether all the parameters being monitored are all correlated, as this is probably when this method is most useful.

In [18], the authors present a method based on minimum entropy set (which is also equivalent to minimum volume set). Given a training data, it tries to find the region with most concentration of data points using e.g. a k-minimum spanning tree, k-nearest neighbour (k-nn) graph, etc. A new data point is tested for anomaly by constructing a new minimum spanning tree. If the data point forms part of this tree or the graph, it is returned as normal else it is flagged as an anomaly. So the advantages of this method include the fact that it does not use a threshold, the result is either yes or no. It is however computationally very expensive as the method involves not only the calculation of distances to the k-nearest neighbours of all the data points but it also involves the weighting of the edge lengths and finally a summation of all the weighted edge lengths for the k-nn groups in the entire set to determine the minimum. To circumvent this, the author proposed a leave one out k-nearest neighbour graph method but this is still computationally expensive except the method is developed in such a way as to allow reuse of previous computation results. [19] However, attempted to solve the computation complexity problem in [18] by first arbitrarily splitting the training dataset into two. It then goes further to calculates a bipartite K-knn graph on each data point in one of the two groups of the training dataset (say, the first group) taking K points at a time, this is described as the set of edges connecting each point in each K in the first group to their nearest k neighbours in the second group. (NB.  k and K are different). The result of each K group is then weighted and calculated as before. The anomaly detection decision is also very similar, a new data point is added to the first group and the bipartite graph is calculated for the new dataset. If the minimum volume K-knn graph includes the new data point the data point is tagged normal, else, it is anomalous. It is important to note that this method greatly reduces the computation complexity of [18] but retains all the advantageous statistical properties of [18]. However, this improvement implicitly assumes that the data points in the two groups are identically distributed and thus half of the group can represent the whole group quite well. This may not be so always. In such a situation, [18] will be a better representative of the dataset than [19] and thus provide better results.

[20] Proposes an anomaly detection scheme for localisation in wireless sensor networks in an indoor environment with a confidence indicator. RSS information is known to be highly degraded in an indoor environment because of multipath and shadowing experienced by the signal. To measure the

spread of the RSS data received; they introduce a median absolute deviation (MAD) from the median, which is then used to calculate the probability density function of the signal at that point. A low MAD score denotes a high confidence indicator, which shows that the data is okay; whereas a high MAD score means a low confidence indicator and the data is flagged as suspicious when it exceeds a threshold. The improvements demonstrated by the authors from their results are however not too high.

In [21] an anomaly detection technique is presented which is slightly different from all other methods considered in that it is approached from a security point of view. This anomaly is believed to be introduced by an enemy for malicious reasons. The enemy wishes to negatively influence decision-making process by misleading the analysers of the received data through replicating sensor nodes. Such anomalies are difficult to detect because the cloned sensor nodes return values, which are 'normal', and as such previous anomaly detection methods will fail to spot them. The solution is then proposed from a security point of view and several methods such as node finger print verification, node geographical information storage, authentication number detection, etc., were presented as solutions. These methods all present ways to identify the genuine member nodes of the wireless sensor network and mark every data from non-member nodes as anomalous. The concerns addressed in the paper are usually not always very important in many applications. It is however worth knowing that an additional layer of security can be added to our sensor data to prevent sabotage.

The method presented in [22] is also similar to [21] in that it presents the detection in terms of a sabotage. The wireless sensor network is designed and each node in the network has accurate knowledge of all its neighbours and their geographic locations. Each node in the network also includes its location in each message it sends. A difference however between [21] and [22] is the fact that the received signal strength of the signal received from each node is modelled to vary based on their geographic location. This information is made available to all the nodes in the network and they use this to estimate the expected RSS they should get from their neighbours. Upon getting something different, it alerts all its neighbours by broadcasting the sensor node that returned the malicious data. The neighbours then evaluate for themselves and determine whether the node is malicious or not, and they respond by telling whether they think it is an anomaly or not. All the responses from other nodes are also broadcasts so they all use information obtained from all their neighbours to update their information accordingly.

Information theoretic methods define anomalies as points which increases the size of the code word required to represent a data set [13]. This is quite understandable since it is easier to refer to a list of same item, than to try to describe a list of different items.

It is key to note that all the methods enumerated present anomaly detection from different perspectives and the choice of which to use will be application-dependent. The method in [15] might be preferable to that in [1] which this project seeks to verify because it is online, also [18] and [19] are computationally more expensive than [1] and this may be undesirable in many applications. The approach in [21] and [22] are quite useful for sabotage situations, but that is not our focus and thus will not be expected to produce better result than [1]. The approach in [12] assumes a distribution for the data and will perform worse than [1] when the data does not follow this model.

## 2.3 Theory

Anomaly detection methods usually rely on the change in a property of the data or system being monitored to effectively function. Many methods rely in one way or the other on the probability distribution function of the data [1] [14] usually in combination with another property. This project as earlier stated, presents an anomaly detection method based on the change (increase) in entropy that occurs when an anomalous data is obtained as presented in [1]. It works based on the fact that outliers lie at distances farther from a certain number $k$ of nearest neighbours than every other data point in the data set. This distance calculation is crux. This method assumes no specific form for the data measured and so the pdf of the data is first calculated. Data split technique is used which involves arbitrarily splitting the measured data into two. The first set is used to calculate the pdf of the data from the following equation:

$$f(X) = \frac{k-1}{MV_k(X)} \qquad (2.1)$$

Where: the first set of the data points has been labelled $M$,

$k$ = number of nearest neighbours as stated above, and

$V_k(X)$ = Volume of the k-nearest neighbour region.

It is worthy of note that $V_k(X)$ depends on the dimension of the data being analysed and will thus have different values for different dimensions. For one dimension, it is simply the length.

The pdf estimate is then plugged-in to the entropy functional, $g(.)$ to determine the entropy functional, $\widetilde{H}$ of the entire data set using the second set of the split data. This functional is finally used to estimate the Shannon entropy, $\widehat{H}$ of the dataset. A threshold is set; if the value of the entropy calculated for a new data set differs from that obtained from that obtained for the training data set by more than the threshold, an anomaly is flagged. Else, the data set is returned as normal.

The equations for $\widetilde{H}$ and $\widetilde{H}$ are as follows;

$$\widetilde{H} = \frac{1}{N} \sum_{i=1}^{N} g(f(X_i)) \qquad (2.2)$$

Where $g(.) = \text{Log}(.)$ and

$$\hat{H} = \tilde{H} + [\text{Log}(k-1) - \varphi(k-1)] \qquad (2.3)$$

Where: $\varphi$ = the digamma function which is evaluated as

$$\varphi(k-1) = \frac{\Gamma'(k-1)}{\Gamma(k-1)} \qquad (2.4)$$

Where also, $\Gamma'(k-1)$ is the derivative of the gamma function $\Gamma(k-1)$.

It is worthy of note that in [1] the results used for this project was a special case of the results obtained. The Shannon entropy is a special form of the Renyi entropy [23] where $\alpha = 1$.

$$I_\alpha = \frac{1}{1-\alpha} \text{Log} \int f^\alpha(x) dx \qquad \alpha \in (0,1) \qquad (2.5)$$

Which is also estimated from;

$$\tilde{I}_\alpha = [\frac{\Gamma(k+(1-\alpha))}{\Gamma(k)}(k-1)^{\alpha-1}]^{-1} \hat{I}_\alpha \qquad (2.6)$$

Where $\hat{I}_\alpha = x^{\alpha-1}$, the choice of functional. The Renyi entropy estimate is then given by

$$\tilde{H}_\alpha = (1-\alpha)^{-1} \text{Log}(\tilde{I}_\alpha) \qquad (2.7)$$

The bias of the estimator has been shown in [22], and presented below.

$$B\left(\hat{G}(f)\right) = c_1 \left(\frac{k}{M}\right)^{1/d} + c_2 \left(\frac{1}{k}\right) + o\left(\frac{1}{k} + (\frac{k}{M})^{\frac{1}{d}}\right) \qquad (2.8)$$

Elaborate proofs are a combination of [24] and [25], shorter ones are found in [1].

## 2.4 The Algorithm

The k-nearest neighbour algorithm for the estimation of entropy of a set of data points follows from [1] without any assumption of a specific form for the data being analysed.

1. Collect data point.
2. There may be need to pre-process the data to remove the effects of thermal variations.
3. Specify k, the number of nearest neighbours to use, this depends on the anticipated number of anomalies suspected to be in the dataset.
4. Calculate the distance of every data point to its k- nearest neighbours.
5. Estimate the pdf of the data.
6. Estimate the entropy.
7. Set a threshold, this depends on the desired confidence interval and false alarm rate.

## 3.0 PRACTICAL WORK
### 3.1 Matlab Code

In the matlab code presented in the appendix, the pdf of a Zero mean Gaussian random variable is estimated using the theory presented above without anomalies. The data split method is compared with using the whole data set. The former showed better result, fitting the data better than the latter. Anomalies were deliberately introduced at specific points in the data and it is shown that the pdf of the dataset shows significant difference in such case. The results gotten for the data split method again, is more preferable to that gotten using the entire dataset.

### 3.2 Elementary Results and Findings

This method was applied to a synthetic data from a Gaussian distribution generated in matlab with zero mean and variance 4, 10000 realizations. A plot of the data without anomalies is shown in *Figure 3.1.*

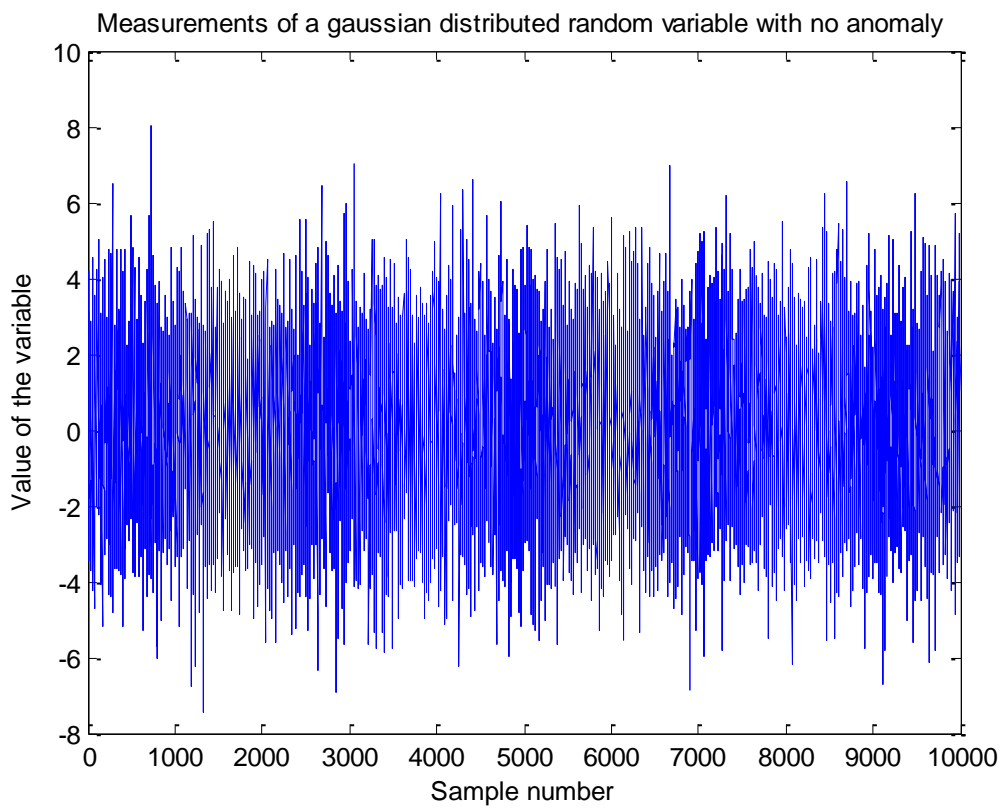

*Figure 3.1:* Gaussian random variable with no anomalous values

The pdf of the data was estimated using the entire dataset of the data set this is as shown in *Figure 3.2*.

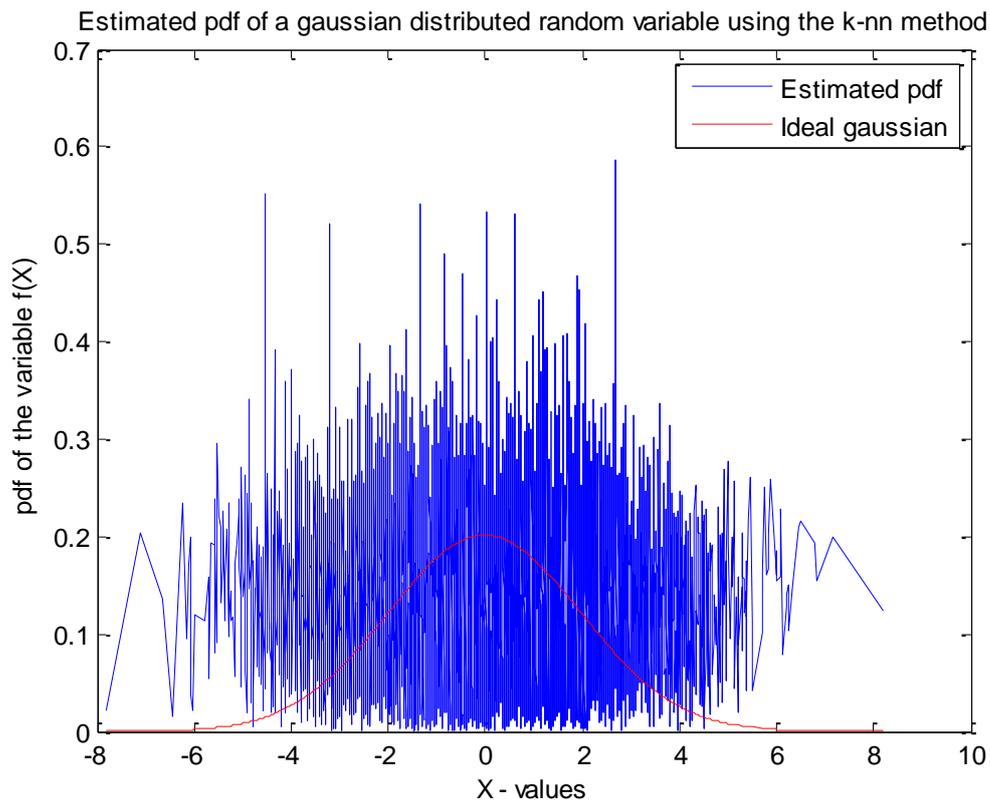

*Figure 3.2:* pdf of the Gaussian random variable estimated using (k =12) k-nn method.

The pdf of the data was estimated using the data split method this is as shown in *Figure 3.3*.

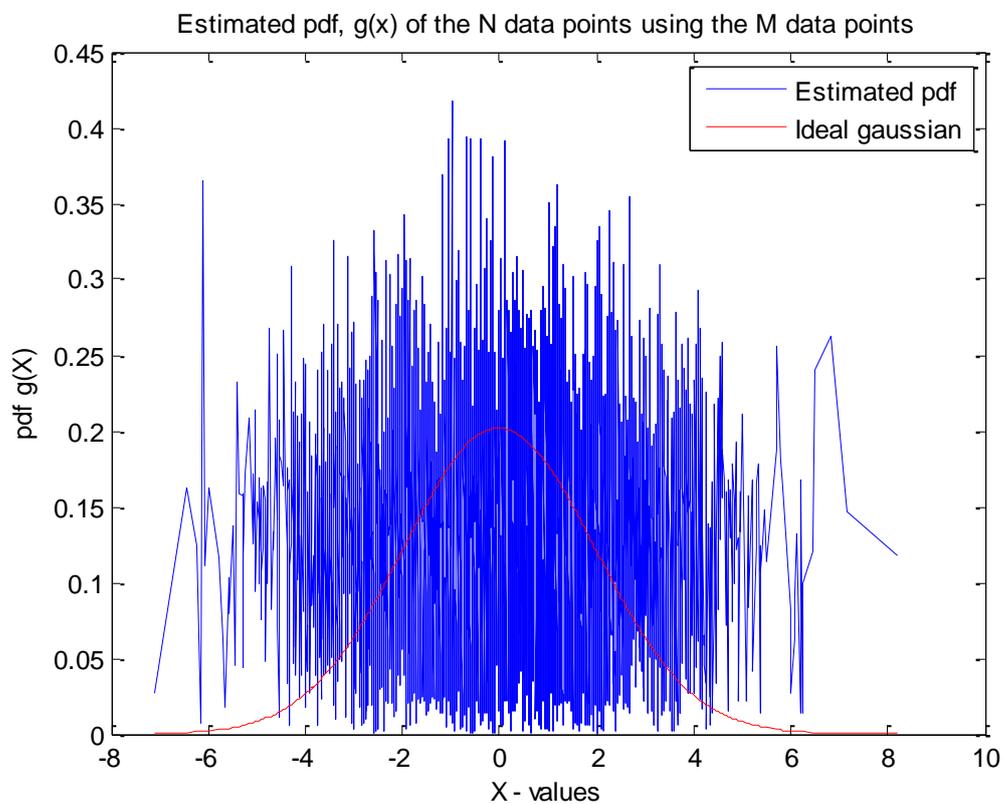

*Figure 3.3:* pdf of the Gaussian random variable estimated using data split (k =12) k-nn method.

Anomalies are intentionally introduced into the data at specific points. A plot of the new data set is shown in *Figure 3.4*.

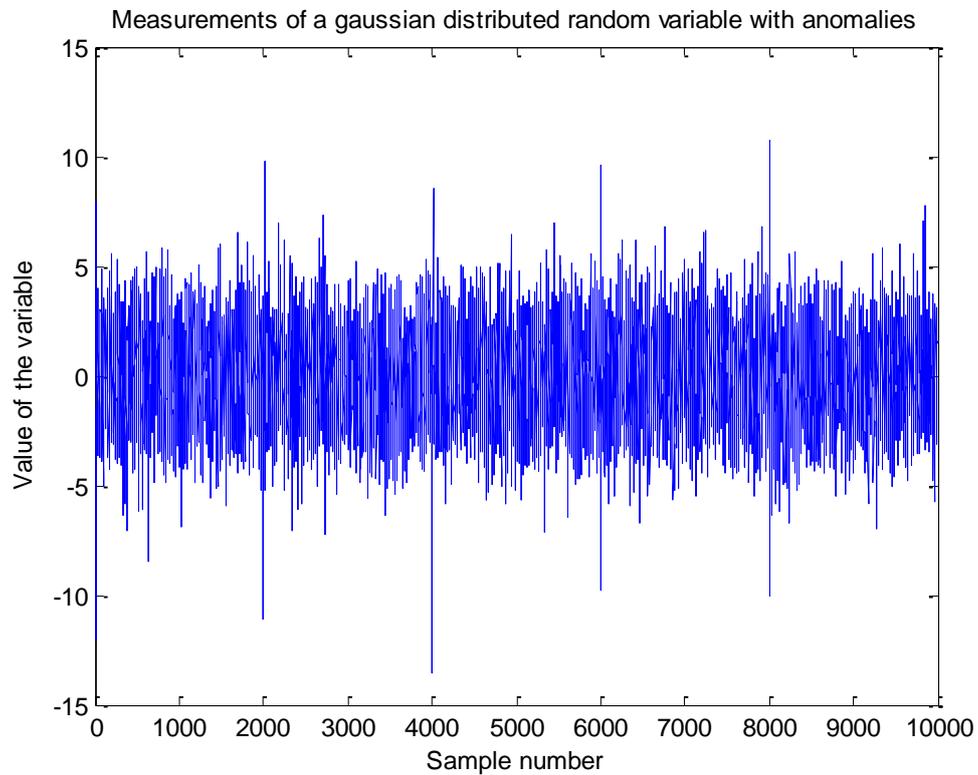

*Figure 3.4:* Gaussian random variable with anomalous values deliberately introduced.

The pdf is then estimated and obtained as shown in *Figure 3.5* using the entire data.

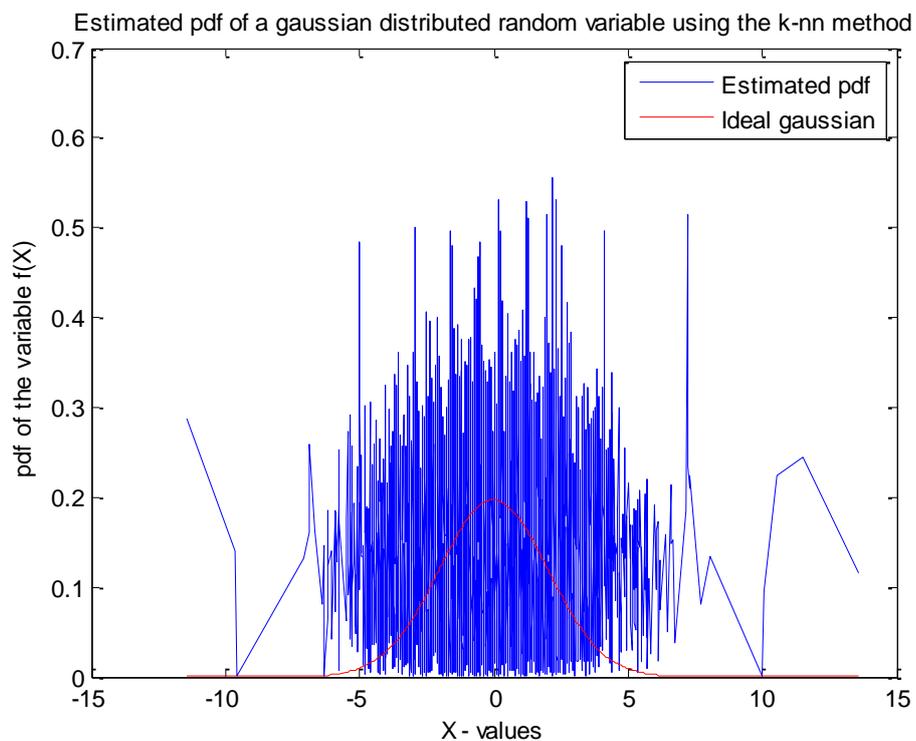

*Figure 3.5:* pdf of the Gaussian random variable with anomalous values deliberately introduced.

Finally, the pdf is estimated using the data split method, it is clear that using data split method provides better results than using the entire dataset.

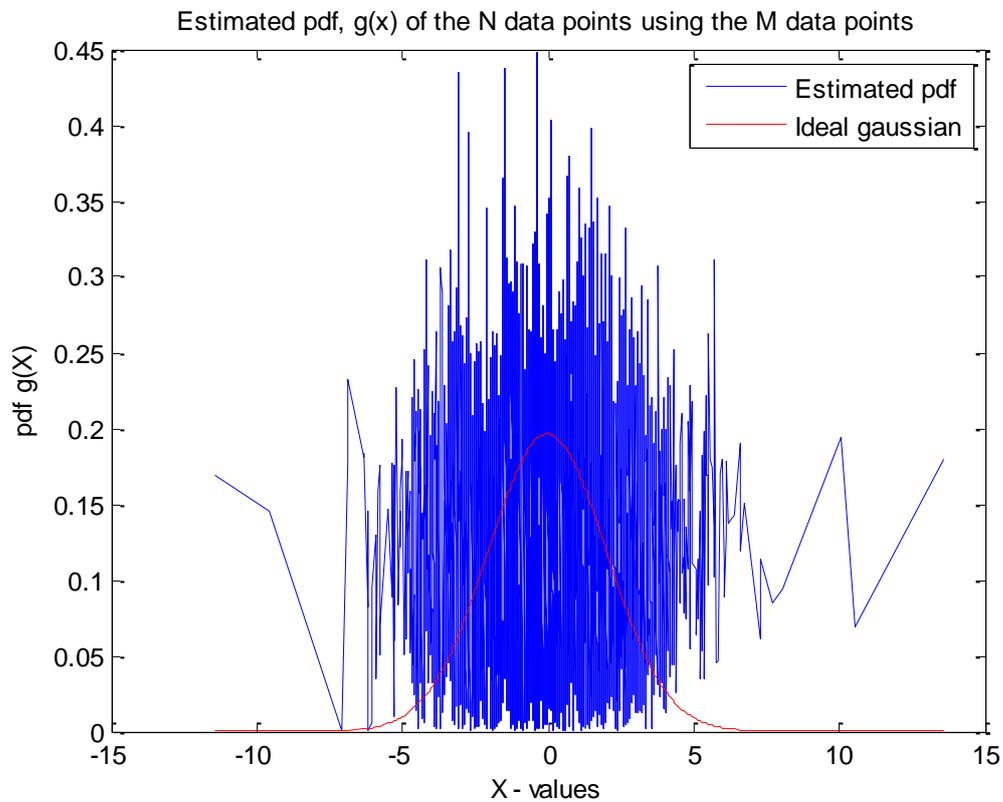

*Figure 3.6:* pdf of the Gaussian random variable estimated using (k =12) k-nn method.

It can be seen from the results presented that the estimate of the pdf obtained for data split method is a better estimate by comparing with that gotten by using the entire dataset using a fitted Gaussian pdf on the same plot as a basis for comparison. Also it is clear that the pdf of the signal changes significantly at anomalous points. This can be easily seen for such points that lie at the tails of the distribution; the pdf exhibits well noticeable haphazard patterns for both the data split method and the full data set method.

## 4.0 PROJECT PLAN AND TIMELINE

### 4.1 Project Plan and Future Work

So far the project has progressed as follows; the literature review presented in this report and preliminary simulations which has been done for the k-nn method using synthetic data (Zero mean Gaussian random variable with variance 4) also presented in this report.

The next phase of the report involves;

1. Evaluation of the k-nn technique on the test signals and comparing the result with the closed form solution for the Gaussian random variable.
2. Detecting anomalies at prior known points.
3. Verification of the technique on the real data.
4. Improving the algorithm.
5. Writing the thesis.

These are structured and planned to take place at the following times;

### 4.1.1   May - June

I expect to fully evaluate the k-nn technique on the synthetic signal presented in this report, estimate its entropy both for the normal case and the anomalous case where anomalies have been included at specific locations, compare this with the closed form for the entropy of a Gaussian equation given as

$$H = \frac{1}{2}\ln(2\pi e\sigma^2) \qquad (3.1)$$

And set a threshold based on this for detecting anomalies with up to 95% confidence interval. To also record the false alarm rates and be sure it corresponds to that reported in [1].

### 4.1.2   June - July

Data has been collected by the authors of [1] as earlier reported and hosted at this website. Successful completion of (1) above will initiate the beginning of verification of the result as in (1) on the real data. The data of course may require some pre-processing to remove the effects of thermal variations of the sensor and its other intrinsic characteristics. This also forms a part of the work that will be done here. Successful verification of the results obtained and presented in [1] will motivate a need to improve the algorithm. The question 'Can this algorithm be better?' will be answered here.

### 4.1.3 July - August

Here, the project work is then presented in the form of a dissertation. It is actual dissertation-writing phase. It is worthy of note to say that this and the above timelines all include a weekly meeting with Professor Mulgrew to evaluate the progress on the project.

### 4.2 Project Dependencies

The different main points in the project timeline presented in the Gantt chart below depend on each other and can be seen to overlap in a lot of cases. For example, progress to section 2 cannot be made without completion of section as this shows that the method is viable and can be used. Events that overlap show actions that can or preferably should be run concurrently.

In section 1, it is expected that the method of anomaly detection presented in this report will be completed for the test signal and results obtained compared with those obtained in [1]. Also closed form solutions exist for the Gaussian case and evaluation will be done against this solutions for the estimated entropy, etc. This runs together with detecting anomalies at known points.

In section 2, in the Gantt chart verification of the results obtained in this project will be done through comparing plots obtained with those presented in [1]. During this process, shortcomings of the method are expected to be vivid and hence a process to improve the algorithm is expected to start.

For section 3, final thesis writing is expected to begin while section 2 is in progress. This is intuitive.

### 4.3 Gantt Chart

The information above is presented in the Gantt chart shown below (which has been divided into two along the middle line for legibility and visibility) to facilitate a visual clarity of the project plan and timeline. Please note that the dates in the chart have been presented in the format MM/DD/YY. The duration is in weeks. Also notes that arrows on the sides of the figures below are meant to show how the lines in each figure correspond to each other.

### 4.4 Expected Challenges and Suggested Solutions

1. Data to be used may contain some irregularities as a result of sensor thermal variations. As mentioned earlier in the report, the project seeks to identify anomalies with the goal of interpretability thus; a pre-processing stage may be required to mitigate these effects.
2. The challenge of interpretability presented above is so as to enable anomalous detection with this method to specify an event, but also a challenge of even being able to identify other forms of anomalies may also prove useful to tackle.
3. Since the actual data to be used may not actually be from a mixture of uniform and beta distributions to which the results are compared, some differences will be expected. But results presented in [1] suggest that this is fairly closely approximated by the method presented.

| Task Name | Start Date | End Date | Duration | Comments |
|---|---|---|---|---|
| Anomaly Detection in wireless sensor networks MSc Project: | 05/25/15 | 08/07/15 | 55 | |
| ⊟ Section 1 - Evaluation using synthetic data | 05/25/15 | 06/15/15 | 16 | |
| 1. Evaluation of the k-nn technique on the test signals | 05/25/15 | 06/05/15 | 10 | How does the result compare with closed form solution 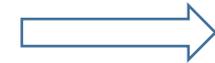 |
| 2. Detecting anomalies at prior known points. | 06/05/15 | 06/11/15 | 5 | Can we detect known anomalies? |
| 3. Comparing plots of results | 06/09/15 | 06/15/15 | 5 | Can we obtain similar plots like in [1]? |
| ⊟ Section 2 - Verification using real data | 06/15/15 | 07/17/15 | 25 | |
| 4. Verification of the technique on the real data | 06/15/15 | 07/10/15 | 20 | |
| 5. Comparing plots of results. | 06/19/15 | 07/13/15 | 17 | Are the results identical with [1]? 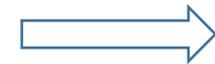 |
| 6. Improving the algorithm. | 06/22/15 | 07/17/15 | 20 | Can the algorithm be better? |
| ⊟ Section 3 - Writing the dissertation | 06/29/15 | 08/07/15 | 30 | |
| 5. Writing the thesis. | 06/29/15 | 08/07/15 | 30 | |

*Figure 4.1a:* Project work plan Gantt chart part 1

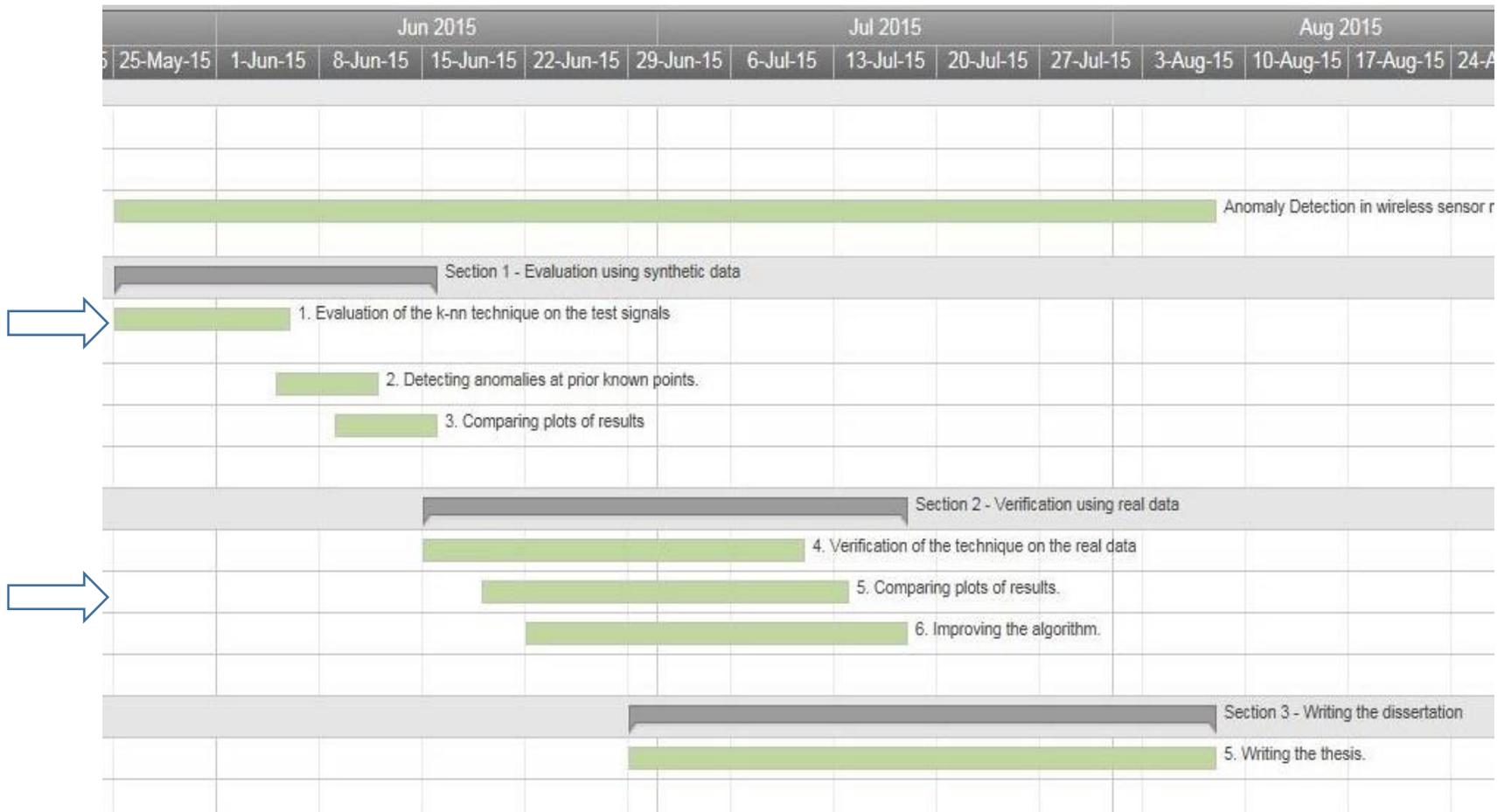

*Figure 4.1b:* Project work plan Gantt chart part 2.

# REFERENCES

[1] K. Sricharan, R. Raich, and A. O. Hero, "k-nearest neighbor estimation of entropies with confidence," in *Information Theory Proceedings (ISIT), 2011 IEEE International Symposium*, aug. 2011, pp. 1205–1209.

[2] Wikipedia. (2015, April 21), Pre dissertation *Wireless sensor network.* [Online]. Available*:* http://en.wikipedia.org/wiki/Wireless_sensor_network

[3] M. Rassam, M. Maarof and A. Zainal, "Adaptive and online data anomaly detection for wireless sensor systems," Elsevier, 2014.

[4] Y Zhang, N. Hamm, N. Meratnia, A. Stein, M. van de Voort and P. Havinga, "Statistics-based outlier detection for wireless sensor networks" in *International Journal of Geographical Information Science*, DOI:10.1080/13658816.2012.654493 27 Feb. 2012, pp2.

[5] S. Siripanadorn, W. Hattagam, and N. Teaumroong, "Anomaly detection using self-organizing map and wavelets in wireless sensor networks," in *Proceedings of the 10th WSEAS International Conference on Applied Computer Science*, ser. ACS'10, 2010, pp. 291–297.

[6] Charu Aggarwal, "An Introduction to Outlier Analysis," in Outlier Analysis, 1st ed., New York, Springer, 2013, ch. i, sec. iii, pp. vi-xvi.

[7] Harshinder Singh, Neeraj Misra , Vladimir Hnizdo , Adam Fedorowicz , Eugene Demchuk, "Nearest Neighbor Estimates of Entropy," *American Journal of Mathematical and Management Sciences*, Vol. 23, Iss. 3-4, 2003

[8] M. N. Goria , N. N. Leonenko , V. V. Mergel , P. L. Novi Inverardi, "A new class of random vector entropy estimators and its applications in testing statistical hypotheses", *Journal of Nonparametric Statistics*, Vol. 17, Issue 3, 2005.

[9] K. Sricharan, R. Raich, and A. Hero. "Empirical estimation of entropy functionals with confidence." *ArXiv e-prints*, December 2010.

[10] Kozachenko, L.; Leonenko, N. "Sample Estimates of entropy of a random vector", *Problems of Information Transmission,* pp23, 95–101, 1987.

[11] Chong Eik Loo , Mun Yong Ng , Christopher Leckie , Marimuthu Palaniswami, "Intrusion Detection for Routing Attacks in Sensor Networks", *International Journal of Distributed Sensor Networks*, Vol. 2, Issue 4, 2006.

[12] Zhu WT, Zhou J, Deng RH, Bao F. "Detecting node replication attacks in wireless sensor networks: a survey." *Journal of Network and Computer Applications* 2012, pp. 1022–34.

[13] O. Salem, Y. Liu, and A. Mehaoua, "Anomaly detection in medical WSNs using enclosing ellipse and chi-square distance," in *Proc. of the 2014 IEEE International Conference on Communications (ICC'14)*, Sidney, Australia, June 2014, pp. 3658–3663.

[14] M. Rassam, A. Zainal, and M. Maarof, "Advancements of Data Anomaly Detection Research in Wireless Sensor Networks: A Survey and Open Issues," *Sensors*, vol. 13, no. 8, pp. 10 087-10 122, 2013.

[15] D. Hawkins, "Identification of Outliers", Chapman and Hall, 1980.

[16] Bhojannawar, S. Satish, M. Chetan, Bulla, and V. Danawade. "Anomaly Detection Techniques for Wireless Sensor Networks-A Survey." *International Journal of Advanced Research in Computer and Communication Engineering* Vol. 2, Issue 10, October 2013.

[17] Josie Hughes, Jize Yan, Kenichi Soga, "Development of Wireless Sensor Network using Bluetooth Low Energy (BLE) for Construction Noise Monitoring", *International Journal on Smart Sensing and Intelligent Systems,* Vol. 8, No. 2, June 2015.

[18] B. Sun, X. Shan, K. Wu, and Y. Xiao, "Anomaly Detection based Secure In-Network Aggregation for Wireless Sensor Networks," *IEEE Systems Journal*, Vol. 7, No. 1, March 2013.

[19] M Xie, S Han, B Tian, S Parvin, "Anomaly detection in wireless sensor networks: a survey", *Journal of Network and Computer Applications*, Vol 34 pp. 1302–1325, 2011.

[20] Y. Chen and J. Juang, "Outlier-Detection-Based Indoor Localization System for Wireless Sensor Networks," International Journal of Navigation and Observation, no. 1–11, 2012.

[21] Guorui Li; Ying Wang, "Sketch Based Anomaly Detection Scheme in Wireless Sensor Networks," Cyber-Enabled Distributed Computing and Knowledge Discovery (CyberC), 2013 International Conference on , vol., no., pp.344,348, 10-12 Oct. 2013

[22] Rajasegarar, S.; Leckie, C.; Palaniswami, M., "Anomaly detection in wireless sensor networks," Wireless Communications, IEEE , vol.15, no.4, pp.34,40, Aug. 2008

[23] Salem, O.; Guerassimov, A.; Mehaoua, A.; Marcus, A.; Furht, B., "Sensor fault and patient anomaly detection and classification in medical wireless sensor networks," Communications (ICC), 2013 IEEE International Conference on , vol., no., pp.4373,4378, 9-13 June 2013

[24] Egilmez, H.E.; Ortega, A., "Spectral anomaly detection using graph-based filtering for wireless sensor networks,"Acoustics, Speech and Signal Processing (ICASSP), 2014 IEEE International Conference on , vol., no., pp.1085,1089, 4-9 May 2014

[25] Jiayu Tang; Pingzhi Fan, "A RSSI-Based Cooperative Anomaly Detection Scheme for Wireless Sensor Networks,"*Wireless Communications, Networking and Mobile Computing, 2007. WiCom 2007. International Conference on* , vol., no., pp.2783,2786, 21-25 Sept. 2007

[26] Bhargava, A.; Raghuvanshi, A.S., "Anomaly Detection in Wireless Sensor Networks Using S-Transform in Combination with SVM," *Computational Intelligence and Communication Networks (CICN), 2013 5th International Conference on* , vol., no., pp.111,116, 27-29 Sept. 2013.

[27] Wikipedia. (2015, August 11),   *Entropy*. [Online]. Available*:* https://en.wikipedia.org/wiki/Entropy_(information_theory)

[28] Anomaly Detection in Wireless Sensor Networks (2015, August 11), [Online]. Available: http://www-personal.umich.edu/~kksreddy/rssdata.html

[29] Afgani, M.; Sinanovic, S.; Haas, H., "Anomaly detection using the Kullback-Leibler divergence metric," *Applied Sciences on Biomedical and Communication Technologies, 2008. ISABEL '08. First International Symposium on* , vol., no., pp.1,5, 25-28 Oct. 2008

[30] Bernard Mulgrew, Peter M. Grant, John Thompson, "Digital Signal Processing: Concept and applications," 2nd ed., Palgrave Macmillan, 2003, ch 9, pp

[31] Sungwook Youn; Chulhee Lee, "Edge Detection for Hyperspectral Images Using the Bhattacharyya Distance," *Parallel and Distributed Systems (ICPADS), 2013 International Conference on* , vol., no., pp.716,719, 15-18 Dec. 2013.

[32] Euisun Choi; Chulhee Lee, "Estimation of classification error based on the Bhattacharyya distance for multimodal data," *Geoscience and Remote Sensing Symposium, 2001. IGARSS '01. IEEE 2001 International* , vol.4, no., pp.1874,1876 vol.4, 2001.

[33] Chulhee Lee; Daesik Hong, "Feature extraction using the Bhattacharyya distance," *Systems, Man, and Cybernetics, 1997. Computational Cybernetics and Simulation., 1997 IEEE International Conference on* , vol.3, no., pp.2147,2150 vol.3, 12-15 Oct 1997.

**APPENDIX**

```matlab
%MATLAB CODE TO ESTIMATE THE PDF AND SHANNON'S ENTROPY
% Code takes in r = measurements or readings(data), k = number of
% neighbours, f = fraction of data split. k = 8; f = 0.6; r = 2*randn(100);

function[H]= myAnomalyDetect(r, k, f)
%set the parameters
%Split the Data

a = r(:);
c = a(1:f*length(a));
d = a(f*length(a)+1:end);
N = f*length(a);
M = (1-f)*length(a);
%initialize the variables
a_distance = zeros(1,M);
a_distance = a_distance(:);
```

```matlab
aj = zeros(size(M+N));
c_distance = zeros(1,N);
c_distance = c_distance(:);
cj = zeros(size(c));
sph = 2; %multiplier for a sphere volume(length) in one-dimension

plot(a);
    title('Measurements of a gaussian distributed random variable with no anomaly')
    xlabel('Sample number') % x-axis label
    ylabel('Value of the variable') % y-axis label
%find nearest neighbors
%k-nn search
    for j = 1 : M+N
        dist = sqrt(( a(j) - a ).^2); %finds the distance to all other points
        ordered = sort(dist); %convert the matrix to a vector and sort it
        a_distance(j) = ordered(k+1);
        aj(j) = a(j);
    end
vol = sph .* a_distance; %value of a sphere volume in one dimension
fx = (k-1)./((M+N)*vol);
a = sort(a);
%fit a gaussian to compare
Ave_d = sum(a)/(M+N);
SD_d = sqrt(var(a));
Y = normpdf(a,Ave_d, SD_d);
%Plot both graphs
figure, plot(a, fx,'b',a,Y,'r');
    title('Estimated pdf of a gaussian distributed random variable using the k-nn method')
    xlabel('X - values') % x-axis label
    ylabel('pdf of the variable f(X)') % y-axis label
    leg_d = legend('Estimated pdf','Ideal gaussian');
    for i = 1:N
        %k-nn search based on the M-points for the N-values
        c_dist = sqrt(( c(i) - d ).^2); %finds the distance to all other points
        c_ordered = sort(c_dist); %convert the matrix to a vector and sort it
        c_distance(i) = c_ordered(k+1);
        cj(i) = c(i);
    end

c_vol = sph .* c_distance; %value of a sphere volume in one dimension
gx = (k-1)./(M*c_vol);
c = sort(c);
%fit a gaussian to compare
    Ave_c = sum(c)/N;
    SD_c = sqrt(var(c));
    X = normpdf(c, Ave_c,SD_c);
    figure, plot(c, gx,'b',c,X,'r');
    title('Estimated pdf, g(x) of the N data points using the M data points')
    xlabel('X - values') % x-axis label
    ylabel('pdf g(X)') % y-axis label
    leg_c = legend('Estimated pdf','Ideal gaussian');

 end
```

Add anomalies

```matlab
%add this snippet immediately after defining 'a' in the code to
%introduce anomalies
    for l = 3:length(a)/5:length(a)
        a(l) = a(l) - 10;
    end
     for l = 10:length(a)/5:length(a)
        a(l) = a(l) + 10;
    end
```